\documentclass{article}

\usepackage[final]{neurips_2023}

\usepackage[utf8]{inputenc} 
\usepackage[T1]{fontenc}    
\usepackage{hyperref}       
\usepackage{url}            
\usepackage{booktabs}       
\usepackage{amsfonts}       
\usepackage{nicefrac}       
\usepackage{microtype}      
\usepackage{xcolor}         
\usepackage{graphicx}
\usepackage{algorithm}
\usepackage{algpseudocode}
\usepackage{makecell}
\usepackage{enumitem}

\title{Effective Quantization for Diffusion Models on CPUs}

\author{%
Hanwen Chang \quad Haihao Shen \quad Yiyang Cai \quad Xinyu Ye \quad Zhenzhong Xu \\
\textbf{Wenhua Cheng} \quad
\textbf{Kaokao lv}  \quad\textbf{Weiwei Zhang} \quad \textbf{Yintong Lu} \quad \textbf{Heng Guo} \quad\\
\texttt{\{hanwen.chang, haihao.shen, yiyang.cai, xinyu.ye, 
zhenzhong.xu}\\
\texttt{wenhua.cheng, kaokao.lv, weiwei.zhang, yintong.lu, heng.guo\}@intel.com}
}

\begin{document}

\maketitle

\begin{abstract}
 Diffusion models have gained popularity for generating images from textual descriptions. Nonetheless, the substantial need for computational resources continues to present a noteworthy challenge, contributing to time-consuming processes. Quantization, a technique employed to compress deep learning models for enhanced efficiency, presents challenges when applied to diffusion models. These models are notably more sensitive to quantization compared to other model types, potentially resulting in a degradation of image quality. In this paper, we introduce a novel approach to quantize the diffusion models by leveraging both quantization-aware training and distillation. Our results show the quantized models can maintain the high image quality while demonstrating the inference efficiency on CPUs. The code is publicly available at: https://github.com/intel/intel-extension-for-transformers.
\end{abstract}

\section{Introduction}

Diffusion models have demonstrated remarkable success in producing images characterized by both high diversity and fidelity, e.g., Stable Diffusion \cite{rombach2022high}, Imagen \cite{saharia2022photorealistic}. Nevertheless, their significant demand for computational resources remains a notable challenge. Although the generated images are undeniably impressive and have captured people's interest, a significant challenge lies in their low performance or high computational costs. Users might find themselves in a situation where generating images on a GPU incurs substantial expenses, and attempting the same task on a CPU results in unacceptably long processing times.

Quantization represents a contemporary area of research aimed at optimizing and improving the efficiency of diffusion methods.
Post-training quantization (PTQ) as outlined in Shang et al.'s research \cite{shang2023post} serves as a valuable reference for applying quantization to diffusion models after the training process. Q-Diffusion\cite{li2023q} divided weights and activations into distinct groups, applying quantization separately to each group. These studies have achieved remarkable Frechet Inception Distance (FID) \cite{heusel2017gans} scores on CIFAR-10, LSUN-Bedrooms, and LSUN-Churches datasets, all while significantly reducing the model's size. While these methods have demonstrated success in terms of FID scores on certain datasets, generating visually appealing images that meet human perception standards remains a persistent challenge.

This paper introduces innovative precision strategies specifically designed for enhancing the performance of Diffusion models. By optimizing performance, we were able to generate images in less than 6 seconds (50 steps) on an Intel CPU, producing output images at a resolution of 512x512 pixels. The image quality has been assessed and confirmed as satisfactory by both human evaluators and FID measurements.
Our contributions can be summarized in three key aspects: 1) Introduce precision strategies/quantization recipes tailored for Diffusion models. 2) Develop an efficient inference runtime equipped with high-performance kernels designed for CPUs. 3) Validate our approach across various versions of Stable Diffusion, including 1.4, 1.5, and 2.1.

\section{Approach}
\label{sec:approach}
\begin{figure}[!htb]
\centering
\includegraphics[scale=0.5]{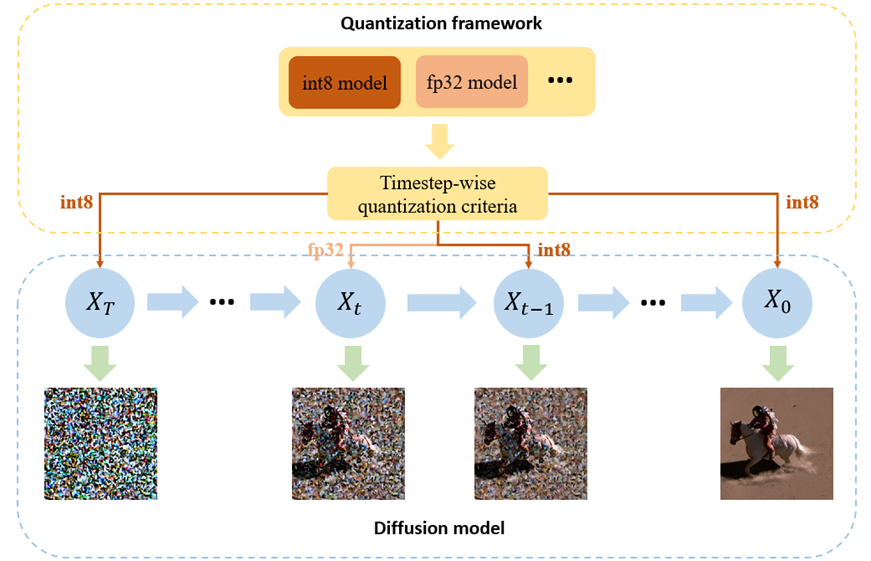}
\caption{Time-dependent Quantization: Different Precisions on Different Steps}\label{Fig6}
\end{figure}

We describe the quantization overview of diffusion models in Figure \ref{Fig6}, which shows the precision is selectively applied per timestamp. 

\subsection{Quantization on Unet}

Numerous diffusion models feature the Unet architecture as a critical element. Throughout the denoising process, the Unet architecture is employed to predict the noise present in the noisy image and subsequently enhance the image by iteratively utilizing this noise estimation across multiple iterations. Profiling analysis reveals that the Unet operation represents the most computationally demanding step in the entire image generation process. As a solution, we have introduced Quantization-Aware Training (QAT) \cite{jacob2018quantization} specifically for the Unet component to alleviate this computational burden. During QAT of Unet, Knowledge Distillation can be incorporated to improve the accuracy. With original Unet as the teacher, its output functions as the guidance for the student, i.e. fake quantized Unet. This quantization workflow is described in Algorithm \ref{alg:qat+kd}.

\begin{algorithm}
\caption{QAT with Knowledge Distillation for Unet}\label{alg:qat+kd}
\begin{algorithmic}
\Require Pretrained diffusion model, dataset, max train steps $N$
\State Copy Unet of pretrained diffusion model as teacher $U_T$;
\State Fake quantize Unet of pretrained diffusion model as student $U_S$;
\For{$k \gets 1$ to $N$}
 \State Sample data from dataset randomly;
 \State Run diffusion model training workflow until Unet's forward;
 \State Get $U_T$'s output $o_T$ and $U_S$'s output $o_S$;
 \State Compute loss between $o_T$ and $o_S$ as $l_{KD}$;
 \State Add loss $l_{KD}$ to the original loss;
 \State Update model's weight with gradient w.r.t. this loss;
 \State Run remaining diffusion model training workflow;
\EndFor
\end{algorithmic}
\end{algorithm}

\subsection{Mixed Precision on Denoising Loop}
The proposed time-dependent mixed precision framework applies mixed precision in a step-wise fashion across the denoising process of the diffusion model. Specifically, within the denoising process spanning 'n' steps, the initial 'k' steps and the final 'k' steps employ a Unet model with higher precision, such as BFloat16, for noise estimation. In contrast, the intervening steps utilize a Unet model with lower precision, like INT8, for noise estimation.

\section{Software Acceleration}
\begin{figure}[!htb]
   \begin{minipage}{0.4\textwidth}
     \centering
     \includegraphics[width=.9\linewidth]{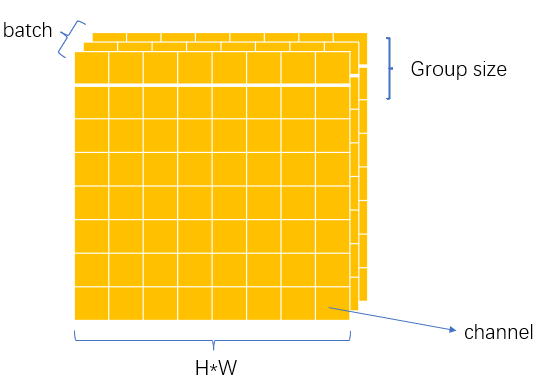}
     \caption{Data Layout}\label{Fig2}
   \end{minipage}\hfill
   \begin{minipage}{0.4\textwidth}
     \centering
     \includegraphics[width=.9\linewidth]{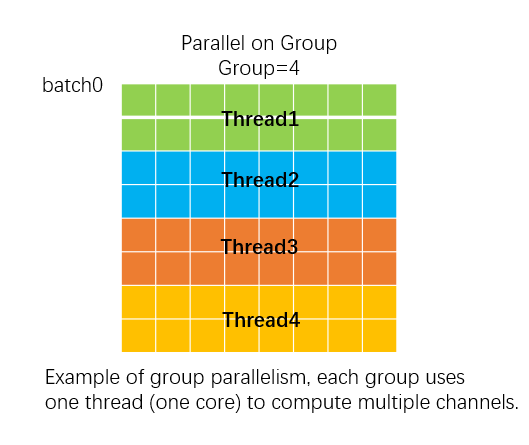}
     \caption{Divide Computing Tasks by Group}\label{Fig3}
   \end{minipage}
\end{figure}

\begin{figure}[!htb]
\centering
\includegraphics[scale=0.4]{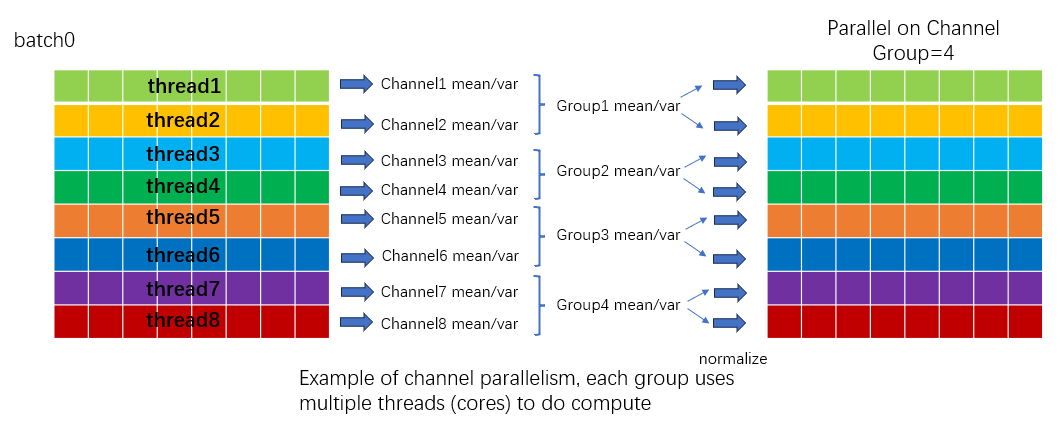}
\caption{Optimized GroupNorm}\label{Fig4}
\end{figure}

While low precision can reduce the inference overhead, we still need to optimize GroupNorm operator. Figure \ref{Fig2} illustrates the data layout, while Figure \ref{Fig3} demonstrates the data division across various cores. The primary issue lies in the fact that the number of groups is fewer than the available CPU cores, resulting in a low CPU utilization rate. To address this issue, we have restructured our approach by computation parallelism across dimensions for channels rather than groups. In the initial step, each core calculates the mean and variance for its respective channels. The subsequent step involves computing group-level values from the channels within each group. Finally, each core performs channel normalization independently. The whole flow is in Figure\ref{Fig4}. You can find these optimizations in Intel Extension for Transformers \cite{itrex}.

Beyond enhancing GroupNorm, we also fuse the Multi-Head Attention (MHA) and introduce an advanced memory allocator to further optimize performance.

\section{Experimental setup}
We select Stable Diffusion as the representative model in our experiment, given it's the most prevalent and widely-used open-source diffusion model. As mentioned in Section~\ref{sec:approach}, we apply quantization to Unet which is performance critical to the entire model. We use the default 50 iterations for latent denoising. Note that there are potential accuracy discrepancies between our model and the others due to configuration differences.

\subsection{Accuracy \& Performance}
On accuracy, we use MS-COCO\cite{Lin2014MicrosoftCC} 2017 validation dataset to evaluate the FID of the Stable Diffusion. The dataset has 5,000 images, and each image has a few captions that describe the image in natural language. We choose 5,000 images and their first caption as the test dataset. Five experimental sets are selected for comparing the FID of Stable Diffusion: \textbf{1.} 50 steps on FP32 Unet; \textbf{2.} 50 steps on BF16 Unet; \textbf{3.} 50 steps on INT8 Unet; \textbf{4.} 6 steps (first and last 3 steps) on BF16 Unet and 44 steps on INT8 Unet, and \textbf{5.} 10 steps (first and last 5 steps) on BF16 Unet and 40 steps on INT8 Unet.

On performance, we leverage Intel Extensions for Transformers \cite{itrex} to measure the performance of various Stable Diffusion versions (1.4, v1.5, and 2.1) on Intel's 4th Generation Xeon Scalable Processors (Sapphire Rapids). The image size 512x512 is used. The code is publicly available at: https://github.com/intel/intel-extension-for-transformers.

\section{Results}
Table\ref{sample-table} shows the accuracy as measured by FID~\cite{fid} using the pre-definedconfigurations. 

\begin{table}[ht]
  \caption{FID of each precision}
  \label{sample-table}
  \centering
  \begin{tabular}{llllll}
    \cmidrule(r){1-6}
    Precision & FP32 & BF16 & INT8 &  BF16 (6 Steps)/INT8 &  BF16 (10 Steps)/INT8 \\
    \midrule
    FID & 30.48  & 30.58   & 35.46  & 31.07 & 30.63\\
    \bottomrule
  \end{tabular}
\end{table}

\begin{figure}[!htb]
\centering
\includegraphics[scale=0.5]{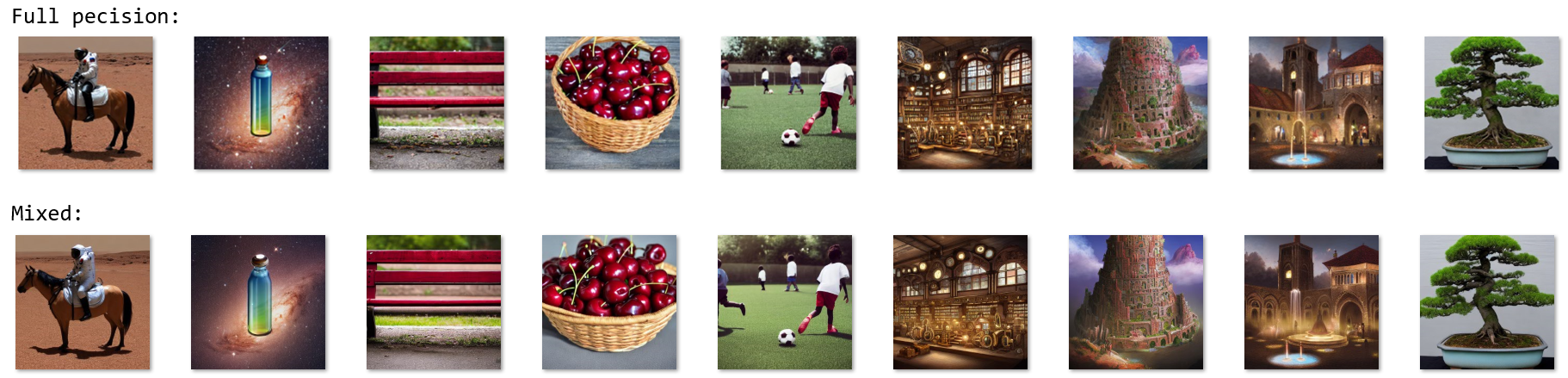}
\caption{output images of mixed precision and full precision.}\label{Fig5}
\end{figure}

You can explore the output images instead of metrics. From Figure~\ref{Fig5}, the image quality looks promising and very close to full precision results. This approach demonstrates its feasibility, with results that are visually indistinguishable to the human eye.

We validated the performance of mixed precision in v1.5, as demonstrated in Table~\ref{perf}, providing compelling evidence that mixed precision can significantly enhance overall performance. In fact, we discovered that employing 20 steps can yield comparable results to using 50 steps. Therefore, we conducted a performance benchmark with the 20-step approach. The latency for BF16 in version 1.5 is 2.74 seconds, while for INT8, it is 2.14 seconds. We hold the belief that a mixed approach could also prove effective. Low precision also works for version 1.4 and version 2.1, their FP32 latency are 11.39 seconds and 16.98 seconds while BF16 latency are both 2.83 seconds.

\begin{table}[ht]
  \caption{Inference Performance (50 Steps)}
  \label{perf}
  \centering
  \begin{tabular}{llll}
    \cmidrule(r){1-4}
    Precision & BF16 &  BF16 (10 Steps)/INT8 & INT8 \\
    \midrule
    Latency &   6.32   & 5.5s  & 5.2s\\
    \bottomrule
  \end{tabular}
\end{table}

\section{Summary and future work}
We presented an effective quantization approach for diffusion models, allowing the mixed precision on Unet to achieve a well-balanced trade-off between accuracy and performance. The next step is to explore other compression techniques such as 4-bits quantization \cite{frantar2022gptq, cheng2023optimize} or sparse \cite{li2023efficient}. We plan to try early exit with INT8 model initialized and subsequently perform inference using mixed precision to improve the quality in https://github.com/intel/neural-speed.

\bibliography{neurips_2023}
\bibliographystyle{abbrvnat}


\end{document}